\documentclass{article}
\usepackage{spconf,amsmath,graphicx}
\usepackage{lipsum}
\usepackage{cite}
\usepackage{amssymb,amsfonts}
\usepackage{algorithmic}
\usepackage{textcomp}
\usepackage{xcolor}
\usepackage{subcaption}
\usepackage{tabularx}
\usepackage{booktabs}
\usepackage{multirow}
\usepackage{graphicx,subcaption,ragged2e}
\usepackage{soul}
\usepackage{bm}
\usepackage{tablefootnote}
\usepackage{hyperref}

\newcommand{\sysname}{AdaNODEs}

\title{AdaNODEs: Test Time Adaptation for Time Series Forecasting Using Neural ODEs}

\name{Ting Dang$^{*}$,  Soumyajit Chatterjee$^{\dagger}$, Hong Jia$^{\ddagger}$, Yu Wu$^{\#}$, Flora Salim$^{+}$, Fahim Kawsar$^{\flat}$ 
\vspace{-1em}
}
\address{
$^{*}$The University of Melbourne, Australia\hspace{0.5em}
$^{\dagger}$Nokia Bell Labs, UK \\
$^{\ddagger}$The University of Auckland, New Zealand\hspace{0.5em}
$^{\#}$University of Cambridge, UK \\
$^{+}$The University of New South Wales, Australia\hspace{0.5em}
$^{\flat}$University of Glasgow, UK
}

\begin{document}
\maketitle
\begin{abstract}
Test time adaptation (TTA) has emerged as a promising solution to adapt pre-trained models to new, unseen data distributions using unlabeled target domain data. However, most TTA methods are designed for independent data, often overlooking the time series data and rarely addressing forecasting tasks. This paper presents \sysname{}, an innovative source-free TTA method tailored explicitly for time series forecasting. By leveraging Neural Ordinary Differential Equations (NODEs), we propose a novel adaptation framework that accommodates the unique characteristics of distribution shifts in time series data. Moreover, we innovatively propose a new loss function to tackle TTA for forecasting tasks. \sysname{} only requires updating limited model parameters, showing effectiveness in capturing temporal dependencies while avoiding significant memory usage. Extensive experiments with one- and high-dimensional data demonstrate that \sysname{} offer relative improvements of 5.88\% and 28.4\% over the SOTA baselines, especially demonstrating robustness across higher severity distribution shifts.
\end{abstract}

\vspace{-5pt}
\section{Introduction}
\vspace{-5pt}
Time series forecasting plays a crucial role in numerous domains. Significant efforts have been dedicated to improving forecasting accuracy, employing techniques from traditional statistical models like autoregressive models~\cite{kaur2023autoregressive} to cutting-edge deep learning methods~\cite{lim2021time}.

While these methods have achieved remarkable success in numerous time series forecasting applications, 
their effectiveness largely depends on the assumption that the training and test data share the same distribution, and data distribution shifts between the training and test sets can cause significant performance degradation~\cite{wang2020tent}.
While domain adaptation techniques, either supervised or unsupervised~\cite{sun2019unsupervised, jin2022domain}, exist to mitigate distributional shifts, they typically require access to a small portion of \emph{labeled} target data~\cite{motiian2017unified} or necessitate access to the original source data. However, in situations where there is no prior knowledge of the incoming data distributions, obtaining labeled target data is costly, or accessing source data raises privacy concerns, these methods often become impractical. These challenges are particularly pronounced when dealing with time series data, as the variations within time series can be pretty diverse, and the labeling process can be exceptionally challenging~\cite{tang2021}. 



Recently, test-time adaptation (TTA) has emerged as a promising solution~\cite{wang2020tent, liang2021source, jia2024tinytta}, which adapts the model using only unlabeled data from the target domain during the runtime and does not need access to source model.
Nonetheless, significant challenges remain: i) most existing TTA methods are primarily designed for independent data types~\cite{wang2020tent, niu2023towards, liang2020we}, rather than for time-series data with temporal dependencies; ii) current TTA techniques are largely focused on classification tasks~\cite{liang2021source,  gong2023sotta, iwasawa2021testtime, dong2025bats}, as they typically depend on optimizing softmax entropy~\cite{wang2020tent}, pseudo-label-based losses~\cite{liang2020we}, or class prototypes~\cite{iwasawa2021testtime}, limiting their applicability to regression tasks, such as time series forecasting.

\begin{figure*}[t]
    \centering
    \includegraphics[width=0.8\textwidth, keepaspectratio]{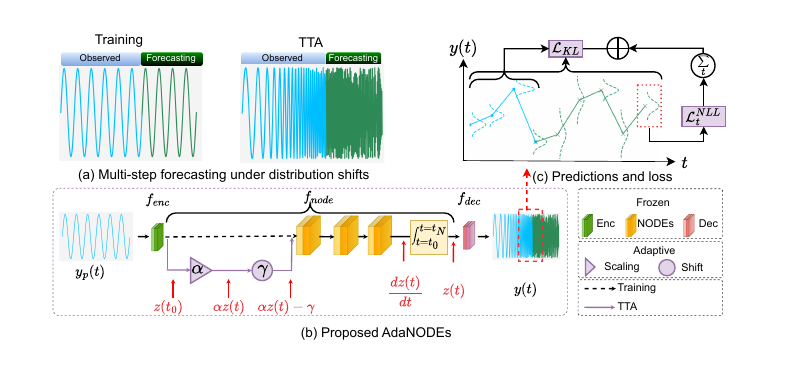}
    \vspace{-10pt}
    \caption{System overview. (a) Distribution shifts in time series forecasting. (b) \sysname{} consists of an encoder, a latent NODEs, and a decoder. During TTA, additional parameters of $\alpha$ and $\gamma$ are incorporated and updated only. (c) The prediction at each time step $t$ is a distribution and the loss function is a combination of the negative log likelihood and KL divergence.} 
    \vspace{-15pt}
    \label{fig:2}
\end{figure*}

To address these challenges, this paper presents \sysname{}, an innovative TTA method for time series forecasting. Specifically, it incorporates two learnable and interpretable parameters into Neural Ordinary Differential Equations (NODEs)~\cite{rubanova2019latent} to adapt to shifts in temporal dynamics within time series data. Additionally, we propose novel loss functions based on negative log-likelihood (NLL) and Kullback-Leibler (KL) divergence, specifically tailored for forecasting tasks. Extensive experiments on both one- and high-dimensional time series data demonstrate the effectiveness of \sysname{}, which opens new pathways for TTA in time series forecasting, and regression in general.

\vspace{-5pt}
\section{Related Work}
\vspace{-5pt}
\label{relatedwork}
\subsection{Test-time Adaptation}
TTA enables a model trained in the source domain to be adapted during runtime without access to the source data and ground-truth labels. 
This, in particular, is significantly advantageous over the commonly used unsupervised domain adaptation~\cite{sun2019unsupervised} and test-time training (TTT)~\cite{sun19ttt} approaches which require access to source data or require a specialized training regime. 
Existing TTA approaches are significantly limited in their ability to model temporal distribution shifts in time series and in their applicability to non-classification tasks, 
like TENT~\cite{wang2020tent}, SHOT~\cite{liang2020we, liang2021source}, 
and SoTTA~\cite{gong2023sotta}, which use entropy minimization (or its variations) and are inherently unsuitable for regression~\cite{zhao2023pitfalls}. Other TTA approaches~\cite{ liang2020we, liang2021source, wang2022continual}, which rely on pseudo-labels for adaptation, are also tailor-made for classification, making them unsuitable for time-series forecasting. 

Recent work has explored TTA for time series, such as TAFAS~\cite{kim2025battling} and PETSA~\cite{medeiros2025accurate}, butt they rely on partial or full labels during inference, which may not be available at test time. SAF~\cite{arik2022self} requires an additional training stage before adaptation, significantly increasing computational cost. Only DynaTTA~\cite{grover2025shift} operates under a similar label-free test-time setting, but it primarily focuses on data normalization and overlooks the inherent temporal dynamics modeling.

\vspace{-5pt}
\subsection{Neural Ordinary Differential Equations}
Neural Ordinary Differential Equations (NODEs) have been introduced to capture dynamic processes using ordinary differential equations (ODEs) whose governing functions are parameterized by neural networks~\cite{rubanova2019latent}. 
Latent ODEs is specifically tailored for time series forecasting~\cite{rubanova2019latent}, which have been widely used in various applications such as health condition forecasting~\cite{ dang2023conditional, dang2023constrained}. NODEs offer flexibility for incorporating constraints into time series modeling~\cite{dang2023conditional, wu2024dual}, enabling the learning of specific temporal patterns and potentially facilitating adaptation to distribution shifts in time series data.
\vspace{-5pt}
\section{Method}
\vspace{-5pt}
\label{methods}
\sysname{} tackles multi-step-ahead forecasting tasks. As shown in Figure~\ref{fig:2}a, the model is trained to predict future values \(y_f(t)\) based on past observations \(y_p(t)\) from time series sampled from the training distribution \(D_{\mathrm{train}}\). At test time, the time series is assumed to come from a different distribution \(D_{\mathrm{test}}\), making it challenging to forecast future steps \(y'_f(t)\) from the context \(y'_p(t)\) due to distributional shifts affecting both past and future segments of the sequence.


\vspace{-5pt}
\subsection{System Overview}
As shown in Figure \ref{fig:2}b, \sysname{} uses a variational encoder-decoder structure consisting i) an encoder $f_{enc}$, ii) latent NODEs $f_{node}$, and iii) a decoder $f_{dec}$. The encoder 
transforms the input into a latent representation \(\mathbf{z}_{t_0}\), capturing past dynamics. Latent NODEs then constructs a trajectory \(\mathbf{z}(t)\) over \([t_0, t_N]\) by solving an ODE with \(\mathbf{z}_{t_0}\) as the starting point. During training, \(\mathbf{z}_{t_0}\) is fed directly to $f_{node}$, while during inference with TTA, additional parameters \(\alpha\) and \(\gamma\) are introduced in $f_{node}$ to adapt to unseen distributions. 
The decoder converts the latent trajectory into the forecasted time series \(y(t)\). As in Figure \ref{fig:2}c, at each time step $t$, the decoder produces a probability distribution over the forecasted values. 

\begin{figure*}[t!]
  \centering
  \subfloat[Amplitude or frequency change]{
    \includegraphics[width=0.47\linewidth, trim=20mm 3mm 10mm 15mm, clip]{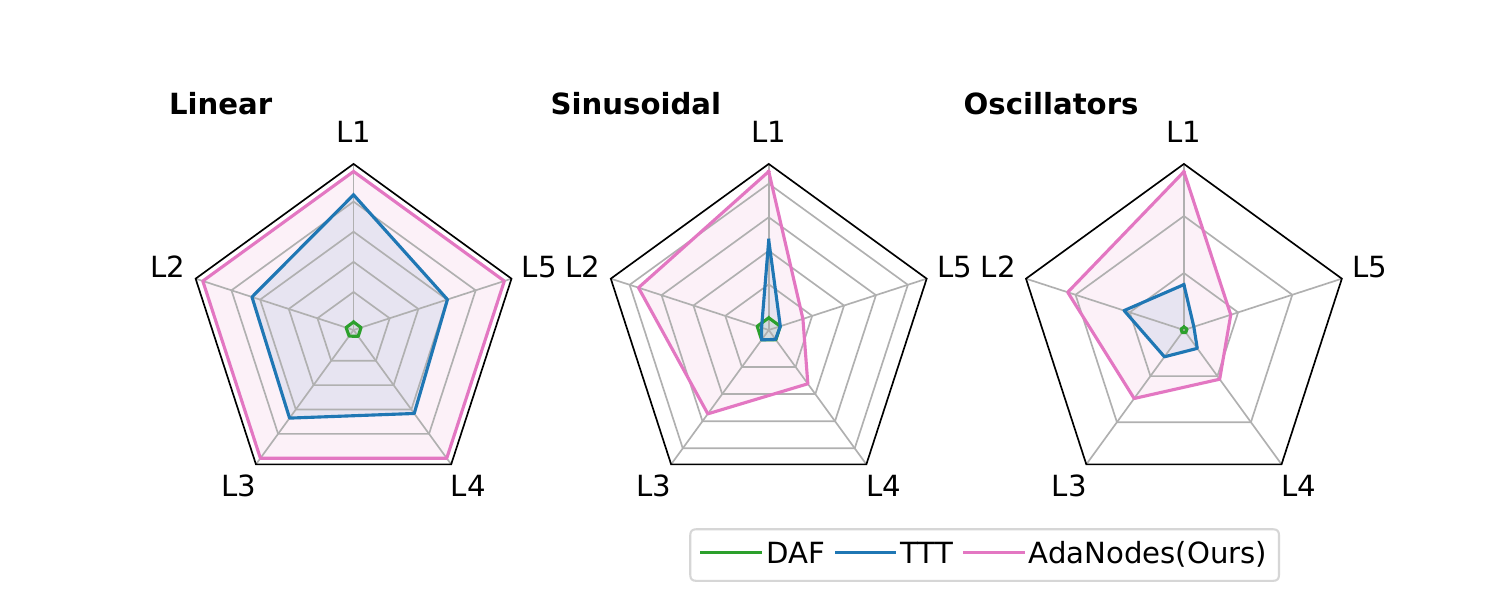}
    \label{fig:radarslope}
  }
  \subfloat[Time delay]{
    \includegraphics[width=0.47\linewidth, trim=20mm 3mm 10mm 15mm, clip]{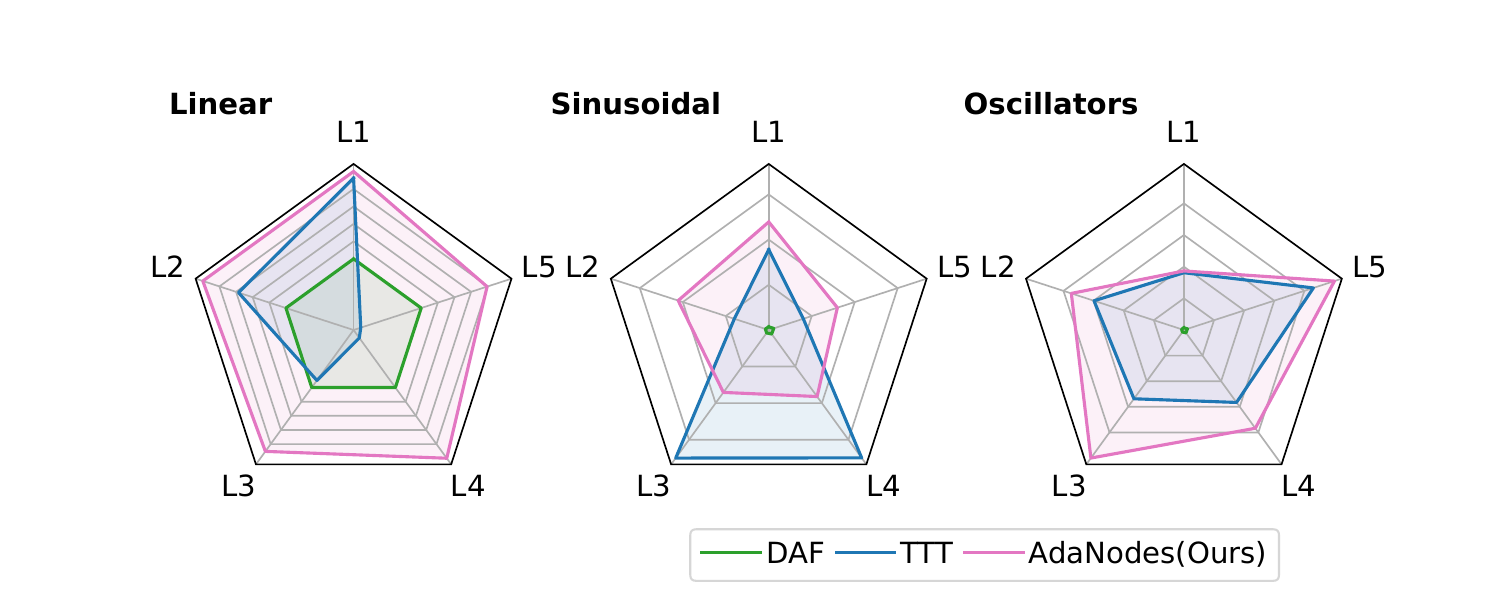}
    \label{fig:radardelay}
  }
  \vspace{-5pt}
  \caption{CCC for one-dimensional signals for severity levels L1-L5 for (a) amplitude or frequency change and (b) time delay.}
  \vspace{-12pt}
  \label{fig:radar}
\end{figure*}

\vspace{-5pt}
\subsection{Adaptation in \sysname{}}
After training the model, we freeze the model parameters and introduce additional parameters \(\alpha\) and \(\gamma\) to accommodate the distribution shifts. Only these two introduced parameters are learnt during the test time. 
The two additional parameters, \(\alpha\) and \(\gamma\), are introduced in latent NODEs as: 
\begin{equation}\label{eq:adanodes}
    \frac{d\bm z(t)}{dt} =  f_{node}(\underline{\alpha} \bm z(t) + \underline{\gamma}; \bm \theta),
\end{equation}
where governing function $f_{node}$ was originally designed to capture the dynamics within the source time series data. By introducing $\alpha$ as a scaling factor and $\gamma$ as a shifting factor, the frequency and phase of the time series in the latent representations $\bm z(t)$ can be adapted. A large $\alpha$ corresponds to higher first derivatives, leading to increased amplitude and a faster rate of change, whereas a small $\alpha$ constrains the derivatives, resulting in lower amplitude and slower change. Meanwhile, $\gamma$ manages the phase shift of the time series, with a positive $\gamma$ inducing a rapid delay and a negative $\gamma$ causing an advance. 
The solution of $\mathbf{z}(t)$ is obtained using the ODE solver:
\begin{equation}\label{eq:5}
    \bm z(t)= \mathrm{ODESolve}(f, \bm \theta, \bm z_{t_0}, t_n, \underline{\alpha}, \underline{\gamma}),\quad
    t_n \in [t_0, t_N]
\end{equation}

Although these two parameters adjusts amplitude/frequency and time shifts linearly, the model is can also handle nonlinearities and complex distribution shifts. This capability arises from the nonlinear mappings provided by the \(f_{node}\) and \(f_{dec}\) functions parameterized by nonlinear neural networks. 

\vspace{-5pt}
\subsection{Loss Function}\label{sec:3.5}
For time series forecasting, the traditional entropy loss function used in TTA for classification is not suitable. \sysname{} introduces a novel loss function. As shown in Figure~\ref{fig:2}c, the prediction at each time step $t$ is modeled as a distribution $p(y_t | \bm z_{t_0}, \bm \theta, \alpha, \gamma)$, where a Gaussian distribution with mean $\mu_t$ and standard deviation $\sigma_t$ is used for tractable solutions. This is expressed as $p(y_t | \bm z_{t_0}, \bm \theta, \alpha, \gamma) \sim N(\mu_t, \sigma_t)$. However, the distribution is flexible and can accommodate various forms.
During the adaptation phase, we adopted an amortized variational inference approach, defined as: 
\begin{align}
    &\underset{\alpha, \gamma}{\text{min}} \mathcal{L}(\bm \theta, \alpha, \gamma) = \lambda\sum_{t}\mathcal{L}_t^{\mathrm{NLL}}(\cdot) +(1-\lambda)\mathcal{L}^{{KL}}(\cdot), \label{eq:loss} \\
    &\mathcal{L}_t^{\mathrm{NLL}} = \mathbb{E}_{\mathbf{z}_{t_0} \sim q_{\bm \phi} (\mathbf{z}_{t_0}|t_\mathbb{C},y_\mathbb{C})} [-\log p(y_\mathbb{T}|\mathbf{z}_{t_0}, t , \alpha, \gamma, \bm \theta)], \label{eq:nll} \\
    &\mathcal{L}^{\mathrm{KL}} 
    = D_{\mathrm{KL}}\Big(q_{\bm \phi} (\mathbf{z}_{t_0}|t_\mathbb{C},y_\mathbb{C}) \;\|\; q_{\bm \phi} (\mathbf{z}_{t_0}|t_\mathbb{T},y_\mathbb{T}) \Big). \label{eq:kl}
\end{align}
\noindent which comprises the negative log-likelihood $\mathcal{L}_t^{NLL}$ and the KL divergence $\mathcal{L}^{KL}$. $\lambda$ is used to balance these two losses. 

As ground truth labels are unavailable, we computed $\mathcal{L}^{NLL}_t$ as the fitting of the predicted mean to the predicted distributions in Eq.~\eqref{eq:nll}. Minimizing $\mathcal{L}^{NLL}_t$ reduces prediction uncertainty, and lower values indicate a better match between the predicted mean and distribution. This is analogous to the entropy minimization loss used in TTA for classification~\cite{wang2020tent}.

$\mathcal{L}^{KL}$ measures the divergence between posterior distributions $q_{\phi}(*)$ derived from context ($*_\mathbb{C}$: past observations) and target ($*_\mathbb{T}$: future predictions) vectors. Minimizing KL divergence aligns predicted and observed distributions, addressing distribution shifts across the entire time series. Thus, the posteriors for observed and entire sequences remain consistent.
\begin{table*}[t!]
\centering
\scriptsize
\caption{Performance comparison on rotating MNIST dataset. In-dist means in-distribution performance.}
\resizebox{0.92\textwidth}{!}{
\begin{tabular}{@{}c|c|ccc|ccc@{}}
\toprule
\textbf{Severity}&\textbf{Models}&\multicolumn{3}{c}{\textbf{Amplitude/Frequency change}}&\multicolumn{3}{c}{\textbf{Time delay}}\\ \midrule
&&\textbf{MSE}$\downarrow$ &\textbf{CC}$\uparrow$ &\textbf{CCC}$\uparrow$ &\textbf{MSE}$\downarrow$ &\textbf{CC}$\uparrow$ &\textbf{CCC}$\uparrow$  \\ \cmidrule{2-8}
    0 &In-dist& 0.023 $\pm$ 0.001 & 0.728 $\pm$ 0.009 & 0.654 $\pm$ 0.011 & 0.023 $\pm$ 0.001 & 0.728 $\pm$ 0.009 & 0.654 $\pm$ 0.011 \\ \midrule
     \multirow{2}{*}{1} & Src&0.034 $\pm$ 0.002 & 0.579 $\pm$ 0.019 &0.526 $\pm$ 0.014 & 0.029 $\pm$ 0.002 & 0.538 $\pm$ 0.021 & 0.475 $\pm$ 0.020 \\ 
     & \sysname{} &\textbf{0.029 $\pm$ 0.001} & \textbf{0.658 $\pm$ 0.015} & \textbf{0.597 $\pm$ 0.012}&  \textbf{0.029 $\pm$ 0.002} & \textbf{0.549 $\pm$ 0.014} & \textbf{0.485 $\pm$ 0.015}\\  \cmidrule{1-8}
       \multirow{2}{*}{2} &  Src & 0.053 $\pm$ 0.003 & 0.359 $\pm$ 0.049 & 0.322 $\pm$ 0.040 &0.042 $\pm$ 0.004 & 0.415 $\pm$ 0.047 & 0.333 $\pm$ 0.038\\ 
     & \sysname{} & \textbf{0.041 $\pm$ 0.002} & \textbf{0.510 $\pm$ 0.029} & \textbf{0.462 $\pm$ 0.024} & \textbf{0.041 $\pm$ 0.003} & \textbf{0.432 $\pm$ 0.022} & \textbf{0.348 $\pm$ 0.017}\\  \cmidrule{1-8}
      \multirow{2}{*}{3} &  Src &0.077 $\pm$ 0.002 & 0.094 $\pm$ 0.036 & 0.084 $\pm$ 0.034& 0.058 $\pm$ 0.003 & 0.205 $\pm$ 0.048 & 0.136 $\pm$ 0.034 \\
     & \sysname{} & \textbf{0.073 $\pm$ 0.003} & \textbf{0.131 $\pm$ 0.016} & \textbf{0.116 $\pm$ 0.016} & \textbf{0.057 $\pm$ 0.003} & \textbf{0.233 $\pm$ 0.034} & \textbf{0.159 $\pm$ 0.025}\\  \cmidrule{1-8}
      \multirow{2}{*}{4} &  Src &0.067 $\pm$ 0.003 & 0.209 $\pm$ 0.084 & 0.189 $\pm$ 0.076 &0.073 $\pm$ 0.002 & 0.008 $\pm$ 0.043 & -0.012 $\pm$ 0.026 \\ 
     & \sysname{} & \textbf{0.066 $\pm$ 0.002} & \textbf{0.236 $\pm$ 0.065} & \textbf{0.213 $\pm$ 0.059} & \textbf{0.071 $\pm$ 0.003} & \textbf{0.030 $\pm$ 0.028} & \textbf{0.004 $\pm$ 0.019}\\  \cmidrule{1-8}
      \multirow{2}{*}{5} &  Src &0.072 $\pm$ 0.006 & 0.146 $\pm$ 0.044 & 0.129 $\pm$ 0.036 &0.083 $\pm$ 0.002 & -0.118 $\pm$ 0.035 & -0.096 $\pm$ 0.020 \\ 
     & \sysname{} & \textbf{0.069 $\pm$ 0.005} & \textbf{0.196 $\pm$ 0.039} & \textbf{0.173 $\pm$ 0.033} & \textbf{0.081 $\pm$ 0.002} & \textbf{-0.099 $\pm$ 0.023} & \textbf{-0.083 $\pm$ 0.015}\\ \bottomrule
\end{tabular}
}
\label{tab:mnist}
\vspace{-8pt}
\end{table*}

\vspace{-10pt}
\section{Experimental setup and results}
\label{experiments}
\vspace{-5pt}


\begin{figure}[t] 
    \centering
    \includegraphics[width=0.43\textwidth]{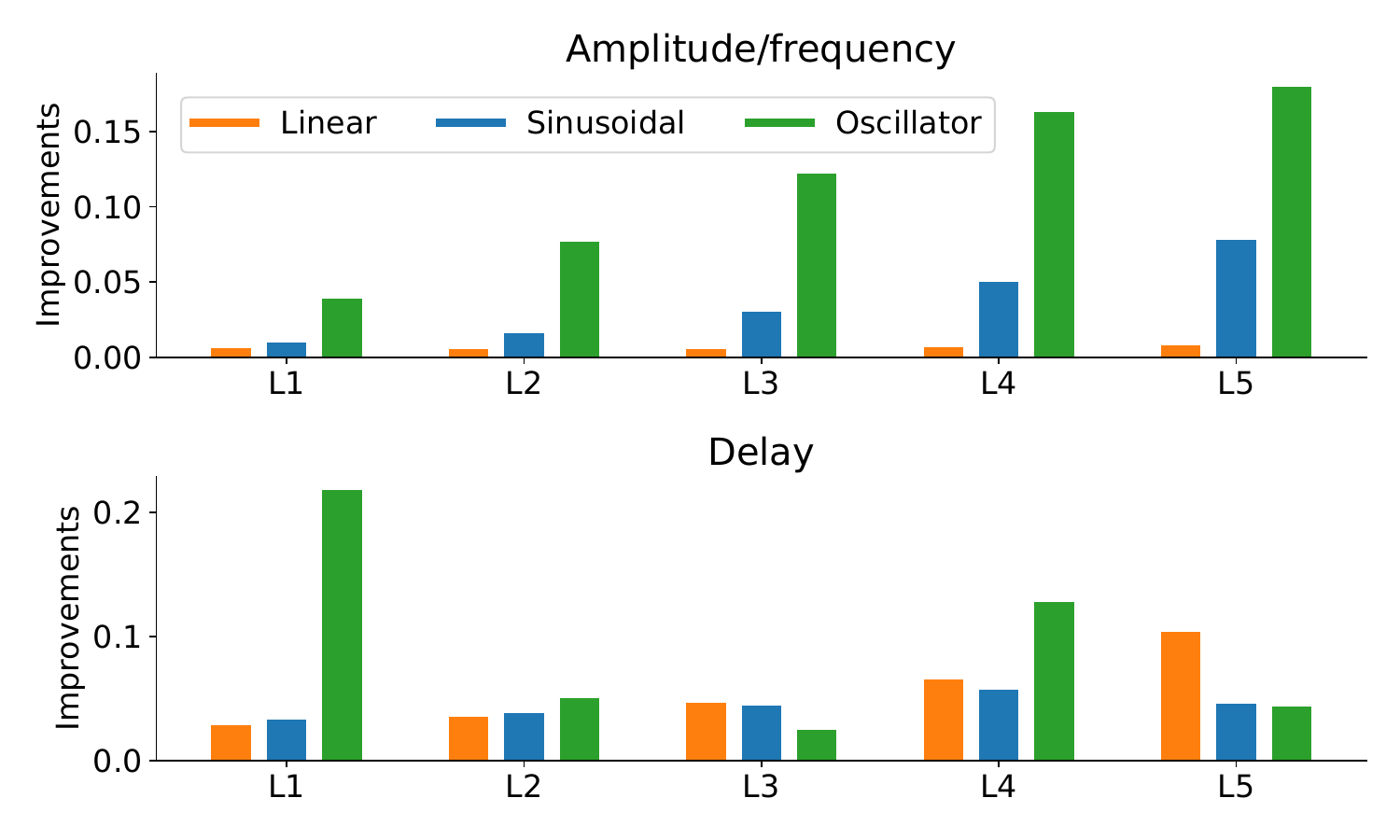}
    \vspace{-5pt}
    \caption{Relative improvements of \sysname{} over source model for all types of signals and severity levels.}
    \vspace{-10pt}
    \label{fig:source}
\end{figure}

\begin{figure*}[ht] 
    \centering
    \subfloat[Severity level 1]{
        \includegraphics[width=0.47\textwidth, trim=3mm 3mm 0 2mm, clip]
        {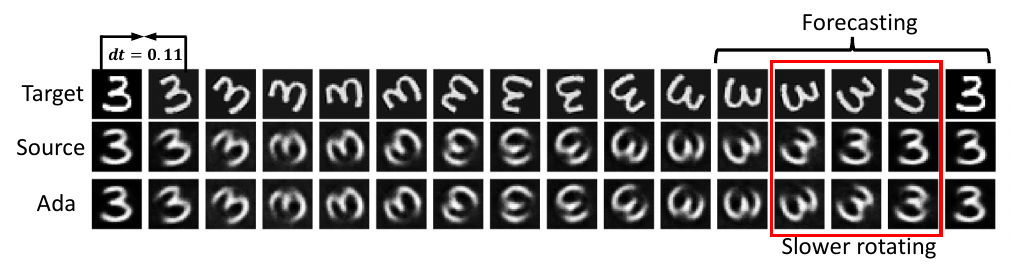}
        \label{fig:mnist1}
    }
    \subfloat[Severity level 5]{
        \includegraphics[width=0.47\textwidth, trim=3mm 3mm 0 2mm, clip]{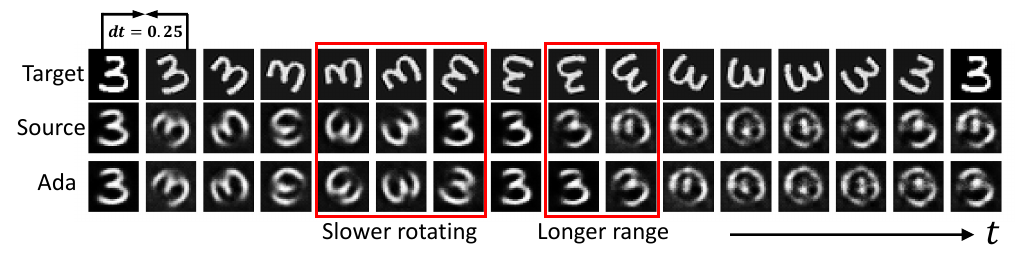}
        \label{fig:mnist5}
    }
    \vspace{-5pt}
    \caption{Rotating MNIST with and without \sysname{} at severity levels 1 and 5. 
    \sysname{} effectively adapts to match the slower rotation speed of the ground truth and extends the prediction over a longer sequence.}
    \vspace{-10pt}
    \label{fig:mnist}
\end{figure*}

\subsection{Experimental Setup} 
\textit{\underline{Data and implementation details.}} We tested \sysname{} on both one-dimensional data and high-dimensional MNIST dataset. We generated one-dimensional linear, sinusoidal, and damped oscillator signals. To induce distribution shifts, we varied amplitude or frequency, and time delays across five severity levels (L1–L5). 
For Rotating MNIST, time series were formed by rotating a digit at a fixed angle, with distribution shifts introduced by changing rotation angles between train and test sets.

For one-dimensional data, we used dense layers for the encoder, NODEs, and decoder. For Rotating MNIST, all layers in \sysname{} utilized 2D CNNs to process images. The model was trained with RMSprop (learning rate optimized within $[10^{-2}, 10^{-3}, 10^{-4}]$), using the adjoint method and RK45 to solve latent NODEs~\cite{rubanova2019latent}. Evaluation metrics include mean squared error (MSE), Pearson's correlation coefficient (CC), and concordance correlation coefficient (CCC)~\cite{lawrence1989concordance}.

\noindent \textit{\underline{Baselines and evaluations.}}
We compared \sysname{} with state-of-the-art (SOTA) models: (1) Source (Src), trained on in-distribution data without adaptation; (2) Unsupervised domain adaptation (DAF)~\cite{jin2022domain}, a SOTA time-series method requiring both source and target data; and (3) Test-time training (TTT)~\cite{sun19ttt}, adapted for regression by using scaling as an auxiliary task and fine-tuning $f_{enc}$ via multitask learning during test time. TTA for classification are not included due to incompatibility~\cite{zhao2023pitfalls}.

\vspace{-5pt}
\subsection{Performance on One-dimensional Data}
For data with amplitude or frequency changes, \sysname{} significantly outperform UDA and TTT across all severity levels and data types as shown in Figure \ref{fig:radarslope}. TTT often performs poorly, even dropping below the source model due to the ad hoc nature of its auxiliary task. DAF shows the weakest performance, optimizing only MSE without predicting discernible trends. 
As severity increases, performance further declines, as the query-key alignment in attention becomes more challenging in DAF. \sysname{} also performs best 
in Figure \ref{fig:radardelay} for shifts in time delays. While TTT shows superior results at sinusoidal severity levels 3 and 4, 
its effectiveness remains sporadic and requires additional training.

We further compared \sysname{} to the source model in terms of relative improvements in Figure \ref{fig:source}. For amplitude or frequency shifts (top), improvements ranged from 0.53\% to 18\% across all signal types, becoming more pronounced at higher severity levels. 
This suggests that \sysname{} effectively discerns underlying temporal patterns especially under severe distortions. For time delays (bottom), \sysname{} consistently outperformed source models, with improvements ranging from 2.5\% to 21.8\%. 

\vspace{-5pt}
\subsection{Rotating MNIST}
We evaluated our model on the rotating MNIST digits dataset, where digit ``3" images rotate over frames. 
Table \ref{tab:mnist} show substantial improvements with \sysname{} over source models. MSE decreased by 9.6\%, while CC and CCC increased by 28.4\% and 28.3\%, respectively, indicating enhanced trend prediction and adaptation to unseen data. Similar improvements were observed for time delays. 

Figures \ref{fig:mnist1} and \ref{fig:mnist5} show rotating MNIST sequences at severity levels 1 ($dt = 0.11$) and 5 ($dt = 0.15$), corresponding to minor and severe corruption, respectively. Here, $dt$ denotes the time interval between consecutive images. Both severity levels have slower rotation speeds than the source sequence ($dt = 0.1$). In both cases, adaptation is needed to slow down the predicted target sequence to match the reduced rotation speed. At severity level 1, the source model predicts faster rotation, but \sysname{} aligns more closely with the target by slowing it down. At severity level 5, \sysname{} successfully slows the rotation to match the extremely slow target speed and extends the forecasting range.

\vspace{-5pt}
\section{Conclusion}
\vspace{-5pt}
In this paper, we proposed \sysname{}, a novel TTA method for time-series forecasting based on NODEs 
to capture the characteristics of distribution shifts in time series data and facilitate effective adaptation during runtime via the novel loss functions. We have demonstrated its efficacy through extensive experiments under various distributional shifts and data types. While the application of TTA to time series forecasting remains a complex task, \sysname{} serves as a promising solution, contributing valuable insights into TTA for time series data and regression tasks in general.



{
\small 
\bibliographystyle{IEEEbib}
\bibliography{refs}

@inproceedings{kim2025battling,
  title={Battling the non-stationarity in time series forecasting via test-time adaptation},
  author={Kim, HyunGi and Kim, Siwon and Mok, Jisoo and Yoon, Sungroh},
  booktitle={Proceedings of the AAAI Conference on Artificial Intelligence},
  volume={39},
  number={17},
  pages={17868--17876},
  year={2025}
}

@article{wang2020tent,
  title={Tent: Fully test-time adaptation by entropy minimization},
  author={Wang, Dequan and Shelhamer, Evan and Liu, Shaoteng and Olshausen, Bruno and Darrell, Trevor},
  journal={International Conference on Learning Representations (ICLR)},
  year={2021}
}

@article{niu2023towards,
  title={Towards stable test-time adaptation in dynamic wild world},
  author={Niu, Shuaicheng and Wu, Jiaxiang and Zhang, Yifan and Wen, Zhiquan and Chen, Yaofo and Zhao, Peilin and Tan, Mingkui},
  journal={International Conference on Learning Representations (ICLR)},
  year={2023}
}

@inproceedings{wang2022continual,
  title={Continual Test-Time Domain Adaptation},
  author={Wang, Qin and Fink, Olga and Van Gool, Luc and Dai, Dengxin},
  booktitle={Proceedings of Conference on Computer Vision and Pattern Recognition},
  year={2022}
}

@inproceedings{liang2020we, 
 title={Do We Really Need to Access the Source Data? Source Hypothesis Transfer for Unsupervised Domain Adaptation}, 
 author={Liang, Jian and Hu, Dapeng and Feng, Jiashi}, 
 booktitle={International Conference on Machine Learning (ICML)},  
 pages={6028--6039},
 year={2020}
}

@article{liang2021source,  
 title={Source Data-absent Unsupervised Domain Adaptation through Hypothesis Transfer and Labeling Transfer}, 
 author={Liang, Jian and Hu, Dapeng and Wang, Yunbo and He, Ran and Feng, Jiashi},   
 journal={IEEE Transactions on Pattern Analysis and Machine Intelligence (TPAMI)},
 year={2021}
}

@article{zhao2023pitfalls,
  title   = {On Pitfalls of Test-Time Adaptation},
  author  = {Hao Zhao and Yuejiang Liu and Alexandre Alahi and Tao Lin},
  journal = {Proceedings of Machine Learning Research},
  year    = {2023},
  pages   = {42058-42080}
}

@inproceedings{sun19ttt,
  Author = {Sun, Yu and Wang, Xiaolong and Zhuang, Liu and
  	Miller, John and Hardt, Moritz and Efros, Alexei A.},
  Title = {Test-Time Training with Self-Supervision for Generalization under Distribution Shifts},
  Booktitle = {ICML},
  Year = {2020}
}

@article{sun2019unsupervised,
  title={Unsupervised domain adaptation through self-supervision},
  author={Sun, Yu and Tzeng, Eric and Darrell, Trevor and Efros, Alexei A},
  journal={arXiv preprint arXiv:1909.11825},
  year={2019}
}

@article{tang2021,
author = {Tang, Chi Ian and Perez-Pozuelo, Ignacio and Spathis, Dimitris and Brage, Soren and Wareham, Nick and Mascolo, Cecilia},
title = {SelfHAR: Improving Human Activity Recognition through Self-Training with Unlabeled Data},
year = {2021},
issue_date = {March 2021},
publisher = {Association for Computing Machinery},
address = {New York, NY, USA},
volume = {5},
number = {1},
journal = {Proc. ACM Interact. Mob. Wearable Ubiquitous Technol.},
month = {mar}
}

@article{gong2023sotta,
  title={SoTTA: Robust Test-Time Adaptation on Noisy Data Streams},
  author={Gong, Taesik and Kim, Yewon and Lee, Taeckyung and Chottananurak, Sorn and Lee, Sung-Ju},
  journal={Advances in Neural Information Processing Systems},
  volume = {36},
  pages={14070-14093},
  year={2023}
}

@article{rubanova2019latent,
  title={Latent ordinary differential equations for irregularly-sampled time series},
  author={Rubanova, Yulia and Chen, Ricky TQ and Duvenaud, David K},
  journal={Advances in neural information processing systems},
  volume={32},
  year={2019}
}

@inproceedings{dang2023conditional,
  title={Conditional Neural ODE Processes for Individual Disease Progression Forecasting: A Case Study on COVID-19},
  author={Dang, Ting and Han, Jing and Xia, Tong and Bondareva, Erika and Siegele-Brown, Chlo{\"e} and Chauhan, Jagmohan and Grammenos, Andreas and Spathis, Dimitris and Cicuta, Pietro and Mascolo, Cecilia},
  booktitle = {ACM SIGKDD conference on knowledge discovery and data mining},
  year={2023},
  pages = {3914-3925}
}

@article{lawrence1989concordance,
  title={A concordance correlation coefficient to evaluate reproducibility},
  author={Lawrence, I and Lin, Kuei},
  journal={Biometrics},
  pages={255--268},
  year={1989},
  publisher={JSTOR}
}

@inproceedings{jin2022domain,
  title={Domain adaptation for time series forecasting via attention sharing},
  author={Jin, Xiaoyong and Park, Youngsuk and Maddix, Danielle and Wang, Hao and Wang, Yuyang},
  booktitle={International Conference on Machine Learning},
  pages={10280--10297},
  year={2022},
  organization={PMLR}
}

@inproceedings{motiian2017unified,
  title={Unified deep supervised domain adaptation and generalization},
  author={Motiian, Saeid and Piccirilli, Marco and Adjeroh, Donald A and Doretto, Gianfranco},
  booktitle={Proceedings of the IEEE international conference on computer vision},
  pages={5715--5725},
  year={2017}
}

@inproceedings{iwasawa2021testtime,
title={Test-Time Classifier Adjustment Module for Model-Agnostic Domain Generalization},
author={Yusuke Iwasawa and Yutaka Matsuo},
booktitle={Advances in Neural Information Processing Systems},
volume = {34},
pages = {2427-2440},
year={2021}
}

@inproceedings{dang2023constrained,
  title={Constrained dynamical neural ode for time series modelling: A case study on continuous emotion prediction},
  author={Dang, Ting and Dimitriadis, Antoni and Wu, Jingyao and Sethu, Vidhyasaharan and Ambikairajah, Eliathamby},
  booktitle={ICASSP 2023-2023 IEEE International Conference on Acoustics, Speech and Signal Processing (ICASSP)},
  pages={1--5},
  year={2023},
  organization={IEEE}
}

@article{wu2024dual,
  title={Dual-constrained dynamical neural odes for ambiguity-aware continuous emotion prediction},
  author={Wu, Jingyao and Dang, Ting and Sethu, Vidhyasaharan and Ambikairajah, Eliathamby},
  journal={Interspeech},
  year={2024},
pages = {3185-3189}
}

@article{jia2024tinytta,
  title={TinyTTA: Efficient Test-time Adaptation via Early-exit Ensembles on Edge Devices},
  author={Jia, Hong and Kwon, Young and Orsino, Alessio and Dang, Ting and Talia, Domenico and Mascolo, Cecilia},
  journal={Advances in Neural Information Processing Systems},
  volume={37},
  pages={43274--43299},
  year={2024}
}

@article{dong2025bats,
  title={E-BATS: Efficient Backpropagation-Free Test-Time Adaptation for Speech Foundation Models},
  author={Dong, Jiaheng and Jia, Hong and Chatterjee, Soumyajit and Ghosh, Abhirup and Bailey, James and Dang, Ting},
  journal={arXiv preprint arXiv:2506.07078},
  year={2025}
}

@article{kaur2023autoregressive,
  title={Autoregressive models in environmental forecasting time series: a theoretical and application review},
  author={Kaur, Jatinder and Parmar, Kulwinder Singh and Singh, Sarbjit},
  journal={Environmental Science and Pollution Research},
  volume={30},
  number={8},
  pages={19617--19641},
  year={2023},
  publisher={Springer}
}

@article{lim2021time,
  title={Time-series forecasting with deep learning: a survey},
  author={Lim, Bryan and Zohren, Stefan},
  journal={Philosophical Transactions of the Royal Society A},
  volume={379},
  number={2194},
  pages={20200209},
  year={2021},
  publisher={The Royal Society Publishing}
}

@article{medeiros2025accurate,
  title={Accurate Parameter-Efficient Test-Time Adaptation for Time Series Forecasting},
  author={Medeiros, Heitor R and Sharifi-Noghabi, Hossein and Oliveira, Gabriel L and Irandoust, Saghar},
  journal={arXiv preprint arXiv:2506.23424},
  year={2025}
}

@article{arik2022self,
  title={Self-adaptive forecasting for improved deep learning on non-stationary time-series},
  author={Arik, Sercan O and Yoder, Nathanael C and Pfister, Tomas},
  journal={arXiv preprint arXiv:2202.02403},
  year={2022}
}

@inproceedings{grover2025shift,
  title={Shift-aware test time adaptation and benchmarking for time-series forecasting},
  author={Grover, Shivam and Etemad, Ali},
  booktitle={Second Workshop on Test-Time Adaptation: Putting Updates to the Test! at ICML 2025},
  year={2025}
}
}

\end{document}